%% file: main.tex
\pdfoutput=1

\documentclass[11pt]{article}

\usepackage[]{acl}

\usepackage{times}
\usepackage{latexsym}

\usepackage{graphicx}
\usepackage{booktabs}
\usepackage{amsmath}
\usepackage{amsfonts}
\usepackage{subfig}
\usepackage{makecell}

\usepackage[T1]{fontenc}

\usepackage[utf8]{inputenc}

\usepackage{microtype}

\usepackage{inconsolata}

\newcommand{\draftonly}[1]{#1} 
\renewcommand{\draftonly}[1]{}

\usepackage{cleveref}
\newcommand{\Scref}[1]{\S\ref{#1}}
\usepackage{enumitem}

\usepackage{tikz}
\newcommand*\circled[1]{\tikz[baseline=(char.base)]{
            \node[shape=circle,draw,inner sep=0.5pt] (char) {#1};}}
            
\usepackage{xspace}
\newcommand{\ours}{\textsc{KeyComp}\xspace} 
\title{Prompting Large Vision-Language Models for Compositional Reasoning}

\author{Timothy Ossowski$^1$, Ming Jiang$^3$, Junjie Hu$^{1,2}$\\
  $^1$Department of Computer Science, $^2$Department of Biostatistics and Medical Informatics\\
  University of Wisconsin, Madison, WI, USA\\
  $^3$ Department of Human-centered Computing, Indiana University, Indianapolis, IN, USA \\
  \texttt{ossowski@wisc.edu}, \texttt{mj200@iu.edu}, \texttt{junjie.hu@wisc.edu}} 

\begin{document}
\maketitle

\input{sections/00_abstract}
\input{sections/01_introduction}
\input{sections/02_method}
\input{sections/03_setting}

\input{sections/04_results}

\input{sections/05_conclusion}

\newpage
\input{sections/06_limitation}

\bibliography{anthology,custom}
\bibliographystyle{acl_natbib}

\input{sections/appendix}

\end{document}

%% file: sections/00_abstract.tex
\begin{abstract}
Vision-language models such as CLIP have shown impressive capabilities in encoding texts and images into aligned embeddings, enabling the retrieval of multimodal data in a shared embedding space. However, these embedding-based models still face challenges in effectively matching images and texts with similar visio-linguistic compositionality, as evidenced by their performance on the recent Winoground dataset. In this paper, we argue that this limitation stems from two factors: the use of single vector representations for complex multimodal data, and the absence of step-by-step reasoning in these embedding-based methods. To address this issue, we make an exploratory step using a novel \textit{generative} method that prompts large vision-language models (e.g., GPT-4) to depict images and perform compositional reasoning. Our method outperforms other embedding-based methods on the Winoground dataset, and obtains further improvement of up to 10\% accuracy when enhanced with the optimal description.\footnote{Code is available at \url{https://github.com/tossowski/KeyComp}.} More importantly, we provide a fine-grained error analysis of our method's outputs, highlighting the key bottleneck in understanding image contents by existing VLMs.
    
\end{abstract}

%% file: sections/01_introduction.tex
\section{Introduction}
\label{sec:introduction}


Recent advancements in vision-language models (VLMs) have rapidly accelerated progress in multimodal tasks such as visual question answering \cite{antol2015vqa} and image captioning \cite{lin2014microsoft}. Large vision-language encoders such as CLIP \cite{radford2021learning} and UNITER \cite{chen2020uniter} have been trained to learn a joint embedding space for combining visual and textual information. These aligned multimodal embeddings have been widely used for zero-shot image-text retrieval \cite{plummer2015flickr30k} and other challenging multimodal matching tasks~\cite{thrush2022winoground}. Notably among these approaches, CACR \cite{pandey2022cross} and IAIS \cite{ren2021learning} further improve the multimodal representations by incentivizing relation alignment during pretraining. 

Despite remarkable advances, the embedding-based methods still encounter difficulties in various compositional reasoning tasks, particularly in the recent Winoground task~\cite{thrush2022winoground}. This task evaluates the capability of VLMs to understand the compositional relation among objects in the images and their associated captions (see details in \Scref{sec:setting}). One primary limitation of embedding-based methods is their reliance on compressing intricate compositional information from an image or a text into a \textit{single vector representation}, which typically emphasizes object encoding but is limited in distinguishing the nuanced relationships between objects and their context in the image and caption. To address this limitation, we propose an \textit{alternative generative} approach that utilizes the fine-grained reasoning capabilities of large generative models in comprehending image content and matching corresponding texts. In contrast to traditional generative methods that train specific modules for visual question answering~\cite{wang2022co, uehara2022learning}, we use a \textit{tuning-free prompt-based} method.




Specifically, our \textit{keyword-guided compositional reasoning} method (\ours) prompts a VLM for depicting images based on the keywords detected from the query text, and then uses a stronger LLM to analyze the image description for matching corresponding texts. Our method design is mainly based on two considerations. First, our analysis (\Scref{sec:results}) shows that directly prompting generative VLMs like MiniGPT4 \cite{zhu2023minigpt} or BLIP-2 \cite{li2023blip} still poses a challenge for the model in identifying key image and text contents for further reasoning. Thus, proper guidance is necessary to instruct the VLM to focus on key image regions for image description. Second, we believe that existing LLMs (e.g., GPT-4) possess stronger language reasoning capabilities than the available VLMs, thus we use a LLM for multi-step reasoning instead of a weaker VLM used in concurrent prompting-based methods~\cite{you2023idealgpt, shen2023hugginggpt, wu2023visual}.

We conduct thorough quantitative and qualitative analyses of our method against existing embedding-based methods on Winoground. Overall, \ours achieves a state-of-the-art image score on Winoground, surpassing the best embedding-based method by a clear margin of 5.1\% image score. Our method excels at multi-step reasoning for complex examples (Fig. \ref{fig:complex_reasoning_text}) and unusual images (Fig.~\ref{fig:non_compositional}). More importantly, further error analysis of the failure cases (Appendix \ref{sec:error_analysis}) reveals a bottleneck in the image description quality of VLMs, shedding insights for future directions. Particularly, VLMs still struggle with describing spatial reasoning and LLMs may misinterpret VLM descriptions with complex syntax structures.




%% file: sections/02_method.tex
\section{Method}

\begin{figure}[t!]%
    \centering
    \includegraphics[width=0.5\textwidth]{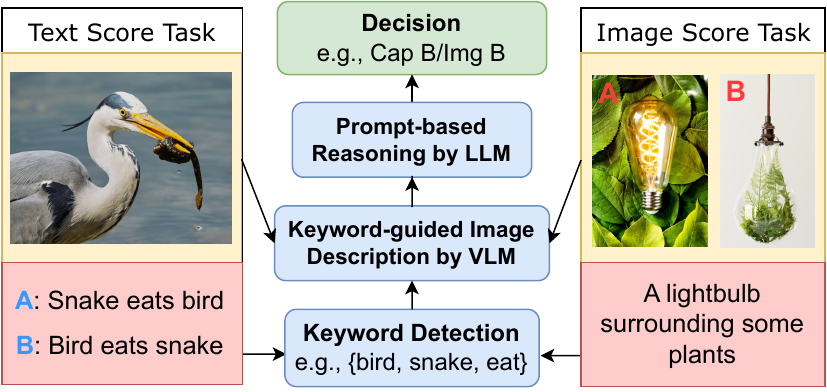}
    \vspace{-7mm}
    \caption{Illustration of our generative method for the Winoground task. Appendix \ref{sec:question_categories} shows more detailed descriptions and model outputs. \textbf{Text Score Task:} Our method chooses the more appropriate caption given a single image. \textbf{Image Score Task:} Our method chooses the best image given a single caption. }
    \label{fig:pipeline}%
    \vspace{-3mm}
\end{figure}
\begin{figure}[t!]%
    \centering
    \vspace{0.5mm}
    \includegraphics[width=0.5\textwidth]{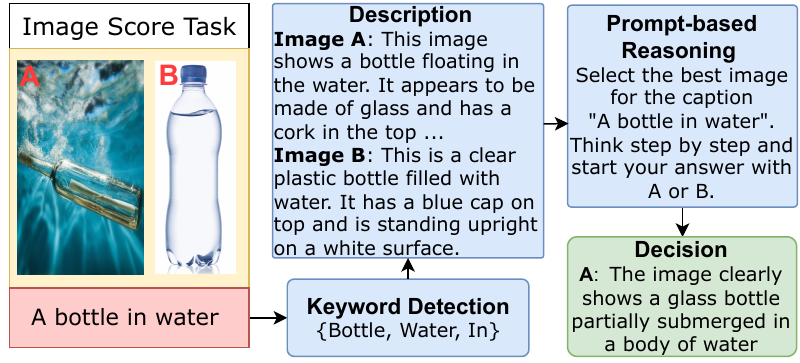}
    \vspace{-7mm}
    \caption{A detailed example for the image score task.}
    \label{fig:image_score_example}%
    \vspace{-3mm}
\end{figure}

This section first describes the Winoground tasks, and then introduces \ours in three steps (illustrated in Fig.~\ref{fig:pipeline} and  Fig.~\ref{fig:image_score_example}). 



\paragraph{Problem Definition} 
Winoground~\cite{thrush2022winoground} has two fundamental matching tasks to assess the visio-linguistic compositionality of VLMs.
Given two aligned image-caption pairs $(I_0, C_0)$ and $(I_1, C_1)$, a VLM is asked to select the correct caption from $\{C_0, C_1\}$ for a single image (either $I_0$ or $I_1$) in the text score task. In the image score task, a VLM is asked to select the correct image from $\{I_0, I_1\}$ for a single caption (either $C_0$ or $C_1$).  

\paragraph{Step 1: Keyword Detection} 
Our method first detects a keyword set $\mathcal{K}=\{w_1,\dots, w_k\}$ automatically from the caption text to help guide the VLM to describe the relevant contents of the images. The caption text is either a single caption in the image task or the concatenation of two captions in the text task. When detecting keywords, we use \texttt{SpaCY}\footnote{\url{https://spacy.io/}} to identify nouns, verbs, prepositions, and adjectives. The intuition is that nouns in a sentence usually describe the key entities in the scene, and adjectives depict the attributes of the entities (e.g., color, size, shape), while verbs and prepositions usually reveal a relation between two entities in an image. 

\paragraph{Step 2: Keyword-guided Description}
Next, we use a pretrained VLM $f_\text{VLM}$ such as MiniGPT4 or BLIP-2 to generate text descriptions for images. To ensure the VLM focuses on relevant image details for image-text matching, we use the keywords $\mathcal{K}$ to guide the generation of image descriptions. Specifically, we append all extracted keywords in $\mathcal{K}$ to the end of a text ``[Instruction]'' to produce a prompt $P_\mathcal{K}$. The prompt is then used to generate a high-quality description of the entities and relations for an image $I_a$ by Eq.~(\ref{eq:img_description}). Example descriptions and prompts are presented in Appendix \ref{sec:question_categories} and \ref{sec:image_score_examples}.
\begin{align}
    P_\mathcal{K} &=\text{``[Instruction] $w_1$, \dots, $w_k$''} \\ \label{eq:img_description}
    D_{\mathcal{K}, a} &= f_\text{VLM}(I_a, P_\mathcal{K}), ~~a \in [0,1]
\end{align}

\paragraph{Step 3: LLM Reasoning \& Explanation}
As LLMs have demonstrated an impressive zero-shot language reasoning ability~\cite{brown2020language}, we prompt a LLM $f_\text{LLM}$ to perform reasoning on the generated image descriptions and the given captions, and \textit{select} an answer for Winoground tasks. Additionally, inspired by recent chain-of-thought prompting~\cite{wei2022chain}, we also add another \textit{explanation} instruction, such as ``Think step by step", to prompt the LLM to explain its answer selection. Specifically, in the image task for selecting the correct image from $I_0, I_1$ for a caption $C_a$, we construct a prompt $P_{\text{img}, a}$ by concatenating $C_a$ with a selection instruction, the two generated image descriptions and an explanation instruction. Similarly, we construct another prompt $P_{\text{txt},a}$ for an image $I_a$ in the text task. Finally, we feed the prompt to the LLM to get a text output $y$ containing the selection and the explanation.
\vspace{-3mm}
\begin{align} 
    P_{\text{img},a} &= \text{``[T] [$C_a$] [T] [$D_{\mathcal{K}, 0}$] [$D_{\mathcal{K}, 1}$] [T]''}\\ 
    P_{\text{txt},a} &= \text{``[T] [$D_{\mathcal{K}, a}$] [T] [$C_0$] [$C_1$] [T]''} \\
    y &= f_\text{LLM}(P_{t,a}), ~~ t \in \{\text{img, txt} \}
\end{align}
where all ``[T]'' placeholders are the texts surrounding the key information in the template to construct the prompt. Appendix \ref{sec:prompt_variations} shows all prompt variants. 

\label{sec:method}

%% file: sections/03_setting.tex
\section{Experimental Settings}

\paragraph{Dataset \& Evaluation}
The Winoground dataset consists of 400 items, each containing two image-caption pairs $(I_0,C_0), (I_1,C_1)$. While the images in each item may be completely different, the two captions $\{C_0,C_1\}$ have an identical set of objects, only in a different order. A model is evaluated by the following text, image, and group scores.
\begin{itemize}[leftmargin=8pt,topsep=1pt] \itemsep-0.2em
    \item \textbf{Text Score}: The model is asked to pick the corresponding caption from $\{C_0,C_1\}$ for a single image $I_a$. The model gets a score of 1 if and only if it picks the correct caption for both $I_0$ and $I_1$.
    \item \textbf{Image Score}: The model is asked to pick the corresponding image from $\{I_0,I_1\}$ for a caption $C_a$. The model gets a score of 1 if and only if it picks the correct image for both $C_0$ and $C_1$.
    \item \textbf{Group Score}: The model achieves a group score of $1$ for the item if and only if it receives a text score of 1 and an image score of 1.
\end{itemize}

\paragraph{Methods in Comparison} We compare our generative method with strong embedding-based methods, i.e., IAIS~\cite{ren2021learning}, CACR~\cite{pandey2022cross} and CLIP~\cite{radford2021learning}. These methods choose the better caption or image by computing a similarity score between their multimodal embeddings and selecting the one with the highest score. In contrast, our generative method generates a text output for selection. We use string matching on the model output to extract the selection for evaluation. If the LLM generates an invalid output (e.g., ``Neither'') indicating neither answer matches the query, we consider the prediction incorrect despite a slight underestimation of our method. Our experiments show that this invalid output occurs about 10\% of the time in the test set. 


\paragraph{Model Selection \& Hyperparameters}
To produce higher-quality descriptions of images, we use an instruction-tuned VLM, i.e., MiniGPT4 \cite{zhu2023minigpt} with a backbone LLM of $\texttt{Vicuna-13b}$. All descriptions are generated using a temperature of 1 and a beam size of 10. With a lower beam size and temperature, we observe notably worse descriptions. For the reasoning component, we utilize OpenAI's ChatGPT API with the $\texttt{GPT-3.5-turbo}$ and $\texttt{GPT-4}$ models using a temperature of 1 and the default values of the other hyperparameters. 

\label{sec:setting}

%% file: sections/04_results.tex
\section{Results and Analysis}
\label{sec:results}
\begin{figure*}[th!]
  \includegraphics[width=\textwidth]{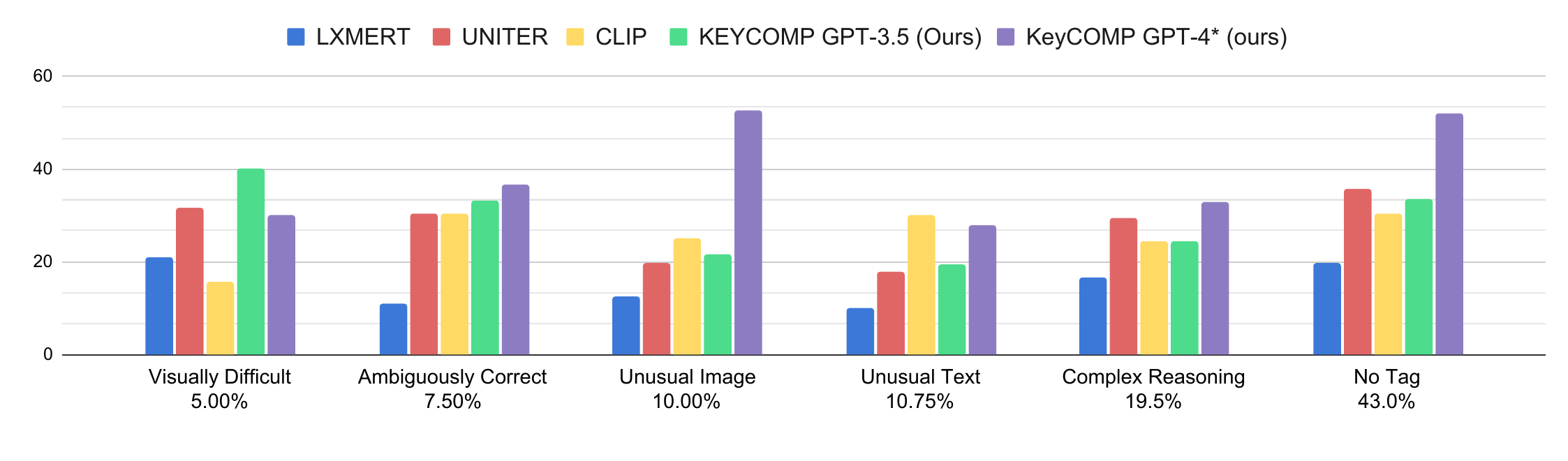}
  \vspace{-13mm}
  \caption{Fine-grained text score performance across different question categories. We give specific examples from each category in Appendix \ref{sec:question_categories}. Percentages on the x-axis indicate each question type's proportion of the dataset. To ensure representative results, question categories comprising less than 5\% of the dataset are excluded.}
  \vspace{-6mm}
  \label{fig:fine_grained}
\end{figure*}

\label{sec:analysis}

\paragraph{Overall Performance}
Table \ref{tab:overall} reports the overall text, image, and group scores of our method compared with existing embedding-based approaches. Since LLMs are generative probabilistic models, they do not have deterministic outputs. Therefore, we report the average score across 3 different runs of the LLM reasoning (Step 3) along with the standard deviation of our method. We observe several findings: (1) Most notably, our method achieves a significantly better image score than prior works, reaching a new state-of-the-art image score. (2) Our text and group scores are also competitive with existing works, even when evaluating accuracy with strict string matching. 
(3) Despite the stochastic nature of our method, the standard deviation of LLM reasoning between runs rarely exceeds $1-2\%$. (4) However, we observe a significant difference in the image description quality from the VLM (Step 2). To estimate an upper bound, we instruct MiniGPT4 to sample 5 descriptions and manually select the best image description to feed into LLMs for reasoning. This leads to a large gain of 12.4\% text score and 3.2\% image score, indicating a potential direction of automatically selecting the best image descriptions for improvement. 

\input{tables/overall.tex}

\paragraph{Image Description Quality Matters.}

Fig. \ref{fig:fine_grained} displays the average text score of our method on various question categories curated by \citet{diwan2022winoground}. We provide correct examples from each category in Appendix \ref{sec:question_categories}, \ref{sec:image_score_examples}, and the image score results in Appendix \ref{sec:fine_grained_text}. Notably, keyword guidance improves image description quality for generative approaches. When providing high-quality image descriptions, our method obtains significant gains in the categories of unusual images (Fig.~\ref{fig:non_compositional}) and complex reasoning (Fig.~\ref{fig:complex_reasoning_text}). For instance in Fig.~\ref{fig:complex_reasoning_text}, when a VLM depicts a rabbit and a turtle correctly, the LLM has the commonsense knowledge to pick the faster one. When prompted with keywords (e.g., ``people'' and ``windows'') in Fig.~\ref{fig:non_compositional}, the VLM is more likely to describe an unusual crayon drawing correctly. This suggests that \ours has the potential to surpass embedding-based methods when external knowledge is necessary for complex reasoning or when precise image information from keywords is required for unusual images. 


\paragraph{Error Analysis and Findings}
We also manually perform fine-grained error analysis on our method's outputs and illustrate three main categories of errors in Appendix \ref{sec:error_analysis}. We identify a bottleneck of image content understanding capability of VLMs and highlight three key findings: 1) the VLM often struggles to describe spatial relationships between two objects, especially for two objects with similar colors (e.g., bushes and hedge in Fig.~\ref{fig:spatial_error}); 2) the LLM occasionally infers the wrong answer due to the misinterpretation of a detailed VLM description with a complicated syntax structure (see Fig.~\ref{fig:llm_error}); 3) the VLM produces inaccurate descriptions of scene elements that are out of focus or missing parts of objects (e.g., a man's head is outside of Fig.~\ref{fig:invisible_error}).

\paragraph{Prompt and Model Ablations}
Table \ref{tab:prompts_avg_std} presents the effect of different prompts on our method. We observe that guiding VLM output with keywords results in the most significant improvement (\circled{2} vs \circled{5}). Tuning the prompts for ChatGPT (\circled{3}-\circled{5}) provides further gains, suggesting future improvement with more sophisticated prompting. Prompting the LLM with chain of thought instructions yields a small gain of 1.3\% text score and 2.3\% image score over a simpler selection prompt (\circled{5} vs \circled{3}). Using a VLM to answer questions directly (\circled{1}) leads to inferior performance, with only 2.0 group score and 11.9\% lower image score than \ours. We also explore different VLM/LLM sizes in Appendix \ref{sec:vlm_llm_size}. 

\input{tables/prompts.tex}

%% file: tables/overall.tex
\begin{table}[thb!]
    \centering
    
    \resizebox{\linewidth}{!} {
    \begin{tabular}{l l l l l}
    \toprule[1pt]
        Method & LLM & Text & Image & Group \\
        \midrule
        \ours$^*$ & GPT-4 & 43.5 $\pm$ 0.7 & 28.7 $\pm$ 2.1 & 18.2 $\pm$ 0.9 \\
        \ours$^*$ & GPT-3.5& 42.7 $\pm$ 0.8 & 27.8 $\pm$  0.7 & 17.4 $\pm$ 0.3 \\
        \ours & GPT-3.5 & 30.3 $\pm$ 1.6 & 24.6 $\pm$ 1.2 & 12.4 $\pm$ 1.2 \\
        IAIS  & - & 42.5 & 19.5 & 16.0 \\ 
        CACR & - & 39.3 & 17.8 & 14.3 \\ 
        CLIP & - & 30.8 & 10.5 & \phantom{0}8.0 \\ 
    \bottomrule[1pt]
   
    \end{tabular}
    }
    \vspace{-2mm}
    \caption{Text, Image, and Group scores scaled by 100 in Winoground. We report the average scores and standard deviation in 3 different runs of LLM reasoning (Step 3). Asterisk indicates an upper bound, where we select the best caption among 5 generated descriptions.}
    \label{tab:overall}
    \vspace{-5mm}
    
\end{table}

%% file: tables/prompts.tex
\begin{table}[thb!]
    \centering
    
    \resizebox{\linewidth}{!} {
    \begin{tabular}{l l l l l}
    \toprule[1pt]
        ID &Prompt Method & Text & Image & Group \\
        \midrule
        \circled{1} & MiniGPT4 end-to-end & $20.8 \pm 0.3$ & $12.5 \pm 1.7$ & \hphantom{0}$2.0 \pm 0.3$ \\
        \circled{2} & No Keyword Guidance & $21.6 \pm 1.0$ & $21.8 \pm 0.8$ & \hphantom{0}$9.4 \pm 1.0$ \\
        \midrule
        \circled{3} & Multiple Choice & $29.0 \pm 0.3$ & $22.3 \pm 0.4$ & $11.9 \pm 0.4$ \\
        \circled{4} &  Explanation Prompting & $26.3 \pm 1.3$ & $23.9 \pm 1.5$ & $11.5 \pm 0.6$ \\
        \circled{5} &   \ours & 30.3 $\pm$ 1.6 & 24.6 $\pm$ 1.2 & 12.4 $\pm$ 1.2 \\
        \midrule
        \circled{6} & Multiple Choice$^*$ & $35.2 \pm 1.2$ & $27.5 \pm 0.7$ & $14.8 \pm 0.7$ \\
        \circled{7} & Explanation Prompting$^*$ & $36.1 \pm 0.4$ & $28.8 \pm  0.6$ & $15.9 \pm 0.6$ \\
        \circled{8} & \ours$^*$ & 42.7 $\pm$ 0.8 & 27.8 $\pm$ 0.7 & 17.4 $\pm$ 0.3 \\
    \bottomrule[1pt]
    \end{tabular}
    }
    \vspace{-3mm}
    \caption{Average performance and standard deviation of prompt variants. To isolate the effect of the prompt, the bottom three rows use the same image description. We show detailed prompt templates in Appendix \ref{sec:prompt_variations}. Asterisk indicates an upper bound, where we select the best caption among 5 generated MiniGPT4 descriptions.
    }
    \vspace{-5mm}
    \label{tab:prompts_avg_std}
\end{table}

%% file: sections/05_conclusion.tex
\section{Discussion and Future Works}
\label{sec:conclusion}

We propose to use VLMs to describe key entities and relations in images to perform compositional reasoning. We empirically demonstrate the effectiveness of our method against embedding-based approaches and end-to-end VLM methods on Winoground. More importantly, we showcase successful examples, categorize common errors made by generative approaches, and identify a key bottleneck of image content understanding of existing VLMs. We hope to shed insights into future works of image content understanding regarding (1) effective prompting strategies to guide VLMs to focus on key image regions; (2) spatial reasoning of objects by VLMs; and (3) accurate interpretation of out-of-focus or partial objects by VLMs. 

%% file: sections/06_limitation.tex
\section*{Limitations}
Our method shows promising results over embedding based methods and end-to-end VLMs. However, small errors in the early steps can accumulate and lead to erroneous reasoning. Using VLMs in a pipeline, as reported by \cite{you2023idealgpt}, is often limited by their performance. This bottleneck explains why we noticed significant improvement when we manually selected the best descriptions generated by MiniGPT4. Additionally, designing a universal prompt for image reasoning can be challenging, and the optimal prompt may change with model updates. Appendix \ref{sec:prompt_variations} illustrates the effect of changing prompts, and future work may consider automatic prompt learning techniques such as prefix tuning \cite{li2021prefix}. Lastly, our method uses a non-deterministic reasoner, resulting in slight output variations even with 0 temperature. Although the variance is low, this may pose issues for downstream tasks.

%% file: sections/appendix.tex
\appendix
\onecolumn
\section*{Appendix}
\section{Related Work}
\paragraph{Vision Language Models}
Language Models and Vision Models by themselves have shown impressive performance in their respective tasks, but combining them to perform vision-language reasoning remains a challenging problem. Popular encoder-based VLMs such as CLIP \cite{radford2021learning} or ALIGN \cite{jia2021scaling} perform contrastive learning on large datasets of image-text pairs, displaying remarkable zero-shot transfer to unseen tasks such as image classification \cite{deng2009imagenet} and image-text retreival \cite{plummer2015flickr30k}. GLIP \cite{li2022grounded} improves the pretraining stage of CLIP by introducing phrase level grounding, allowing for richer semantic representations for the object detection task \cite{ren2015faster}. Recent sequence-to-sequence VLMs such as OFA \cite{wang2022ofa} and FLAVA \cite{singh2022flava} pretrain on a larger variety of vision-language tasks and objectives to allow for more sample-efficient pretraining and cross-task trasnfer. 

 Other recent works combine the information from different modalities by introducing lightweight tunable parameters to connect frozen image and text encoders. Flamingo \cite{alayrac2022flamingo} freezes a vision encoder and inserts trainable cross attention layers to a frozen LLM to integrate visual features, achieving state of the art few-shot performance. BLIP-2 \cite{li2023blip} also freezes a vision encoder and LLM, aligning the modalities with a lightweight transformer called the Q-former. Works such as \textit{Frozen} \cite{tsimpoukelli2021multimodal} and MAPL \cite{manas2022mapl} propose methods which freeze a LLM, but train a visual encoder to represent images as continuous embeddings, resulting in a multimodal few-shot learner. Inspired by the success of instruction tuning LLMs with human feedback with ChatGPT \cite{ouyang2022training}, MiniGPT4 \cite{zhu2023minigpt} and LLaVA \cite{liu2023llava, liu2023improvedllava} further improves the quality of BLIP-2 outputs and LLaMA-2 by instruction tuning with additional descriptions generated by ChatGPT.

\paragraph{Large Language Models as Reasoners}
Concurrent work has explored the idea of using large language models to connect visual foundation models. For example, HuggingGPT \cite{shen2023hugginggpt}, CHAMELEON \cite{lu2023chameleon}, MM-React \cite{yang2023mm}, and Visual ChatGPT \cite{wu2023visual} utilize ChatGPT as a controller, allowing it to delegate visually demanding tasks to foundational visual models. IdealGPT \cite{you2023idealgpt} proposes an iterative approach to decompose complex visual tasks into a series of sub-questions and answers. ChatCaptioner \cite{chen2023video} has BLIP-2 interact conversationally with ChatGPT to create more informative image descriptions. ViperGPT \cite{suris2023vipergpt}, VisProg, \cite{gupta2023visual}, and CodeVQA \cite{subramanian2023modular} use code LLMs such as CodeX \cite{chen2021evaluating} to write python programs which invoke VLMs to solve challenging visual tasks without any task-specific training. Different from these works, we replace the ChatGPT controller with keyword-guided detailed descriptions of the contents of an image, delegating reasoning to a more powerful LLM instead of a VLM. We also avoid iteratively prompting our models to avoid potential error accumulation.
\label{sec:related}
\section{Prompt Variations}
In Tables \ref{tab:prompts} and 
\ref{tab:prompts_image} we display all the different prompt variations used in our experiments.

\input{tables/prompt_table_text}

\input{tables/prompt_table_image_score}
\clearpage
\label{sec:prompt_variations}

\section{Text Score Task for Question Categories}
We compile several examples of our method, with one example from each question category defined by a recent study~\cite{diwan2022winoground}.
\begin{figure}[!htb]
  \includegraphics[width=\textwidth]{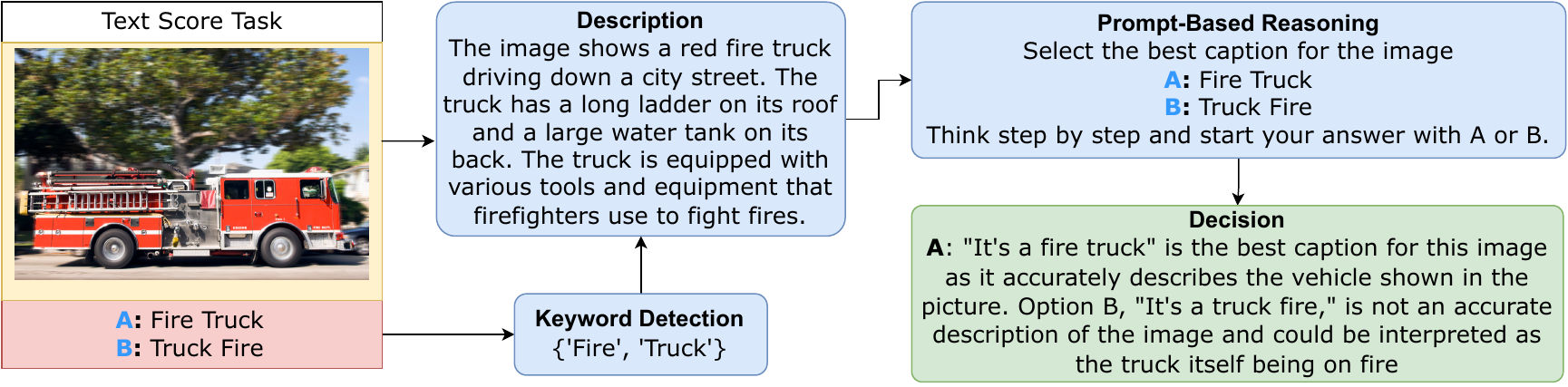}
  \caption{Non-Compositional Question. The swapped words (``Fire'' and ``Truck'') do not necessarily contain the same semantic entities, so compositional reasoning may not be required to solve the question. }
  \label{fig:non_compositional}
\end{figure}

\begin{figure}[!htb]
  \includegraphics[width=\textwidth]{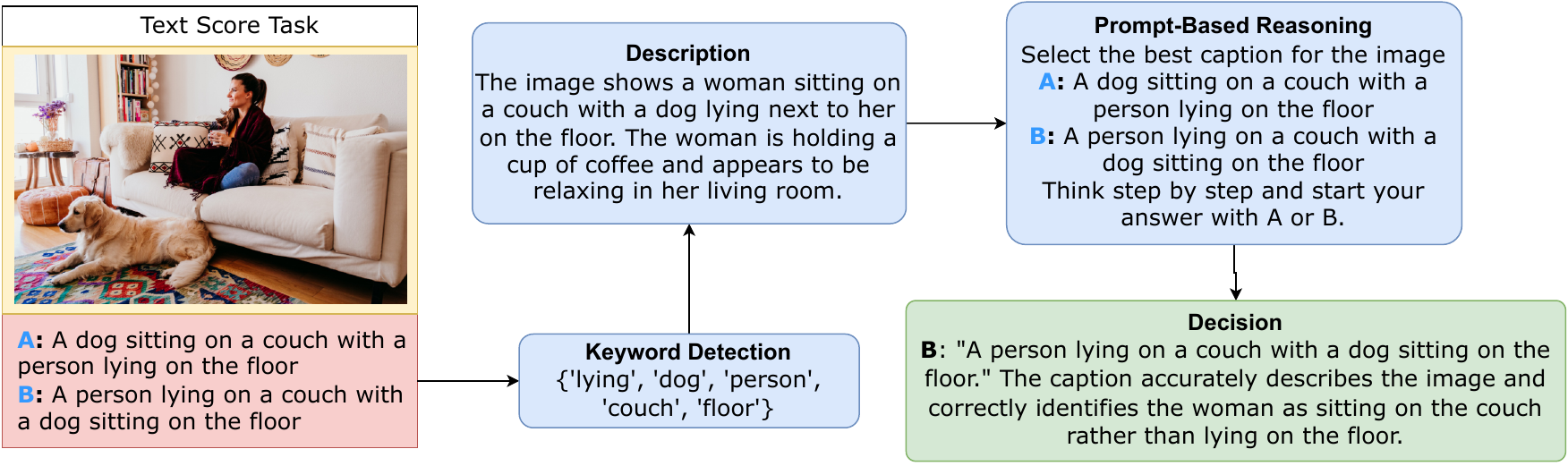}
  \caption{Ambiguously Correct Question. Note that the correct caption B describes the woman as lying on the couch when she is sitting, but the LLM is still able to pick the ambiguously correct caption.}
  \label{fig:non_compositional}
\end{figure}

\begin{figure}[!htb]
  \includegraphics[width=\textwidth]{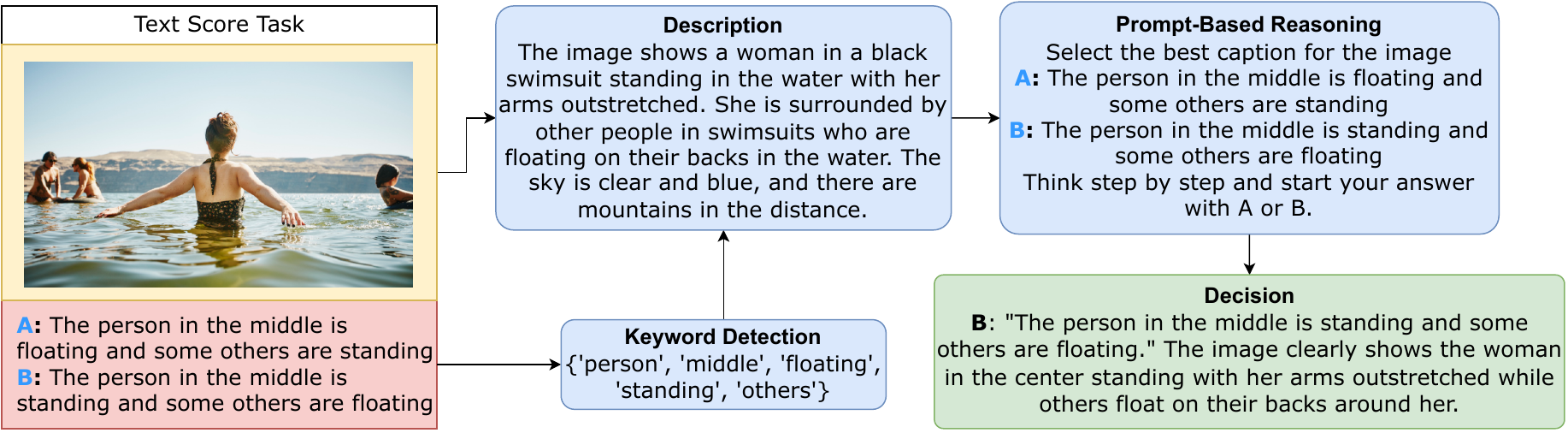}
  \caption{Visually Difficult Question. To correctly answer, the model needs to identify the person floating on the right. }
  \label{fig:non_compositional}
\end{figure}

\begin{figure}[!htb]
  \includegraphics[width=\textwidth]{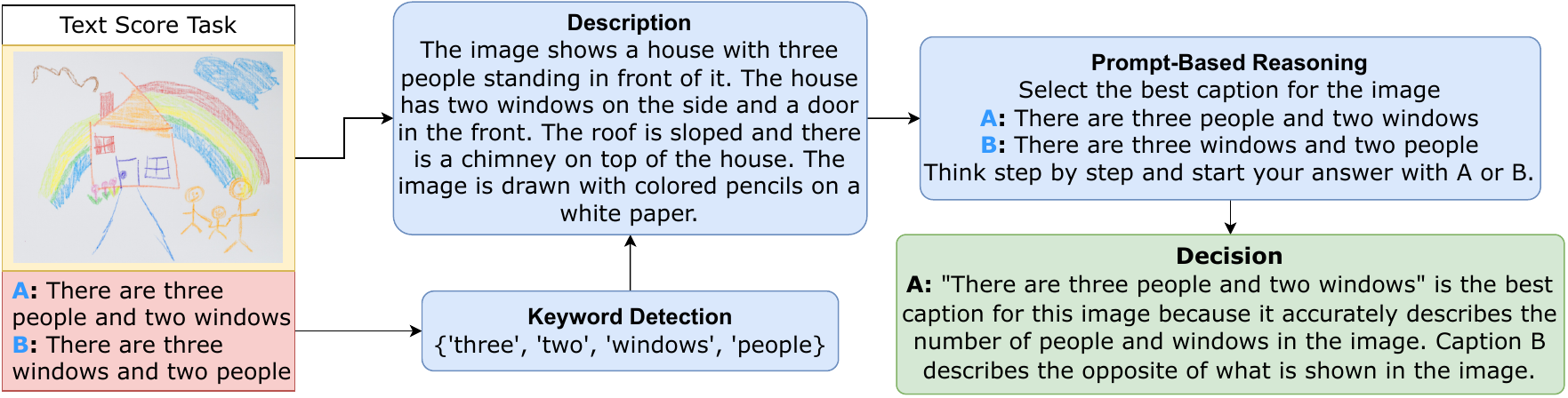}
  \caption{Unusual Image Question. The VLM is more likely to describe an unusual crayon drawing correctly, based on the detected keywords ``windows'' and ``people'' in the prompt, while embedding-based methods are not feasible to include any hints, and are sensitive to domain shifts in the image distribution. }
  \label{fig:non_compositional}
\end{figure}

\begin{figure}[!htb]
  \includegraphics[width=\textwidth]{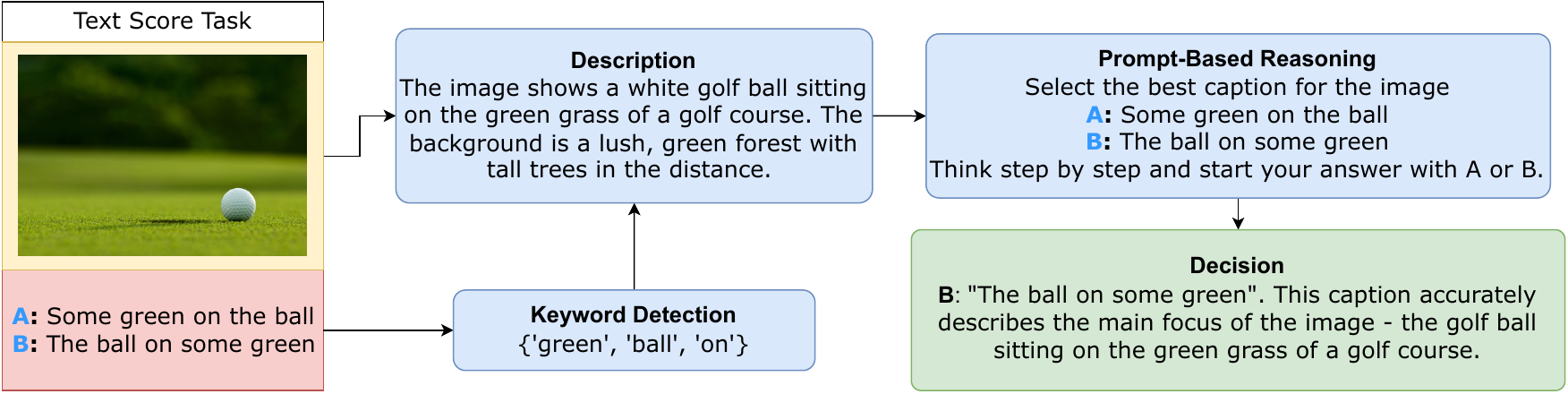}
  \caption{Unusual Text Question. The caption ``The ball on some green" is considered less common compared to an alternative such as ``the ball on some grass".  }
  \label{fig:non_compositional}
\end{figure}

\begin{figure}[!htb]
  \includegraphics[width=\textwidth]{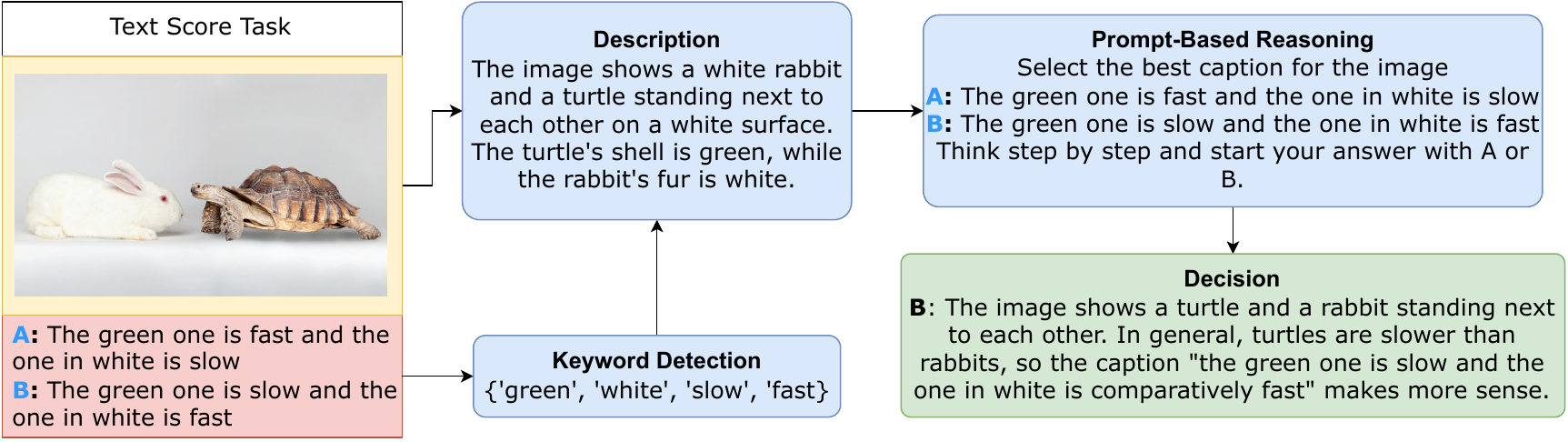}
  \caption{Complex Reasoning Question. This requires the knowledge that rabbits are typically faster than turtles. }
  \label{fig:complex_reasoning_text}
\end{figure}

\begin{figure}[!htb]
  \includegraphics[width=\textwidth]{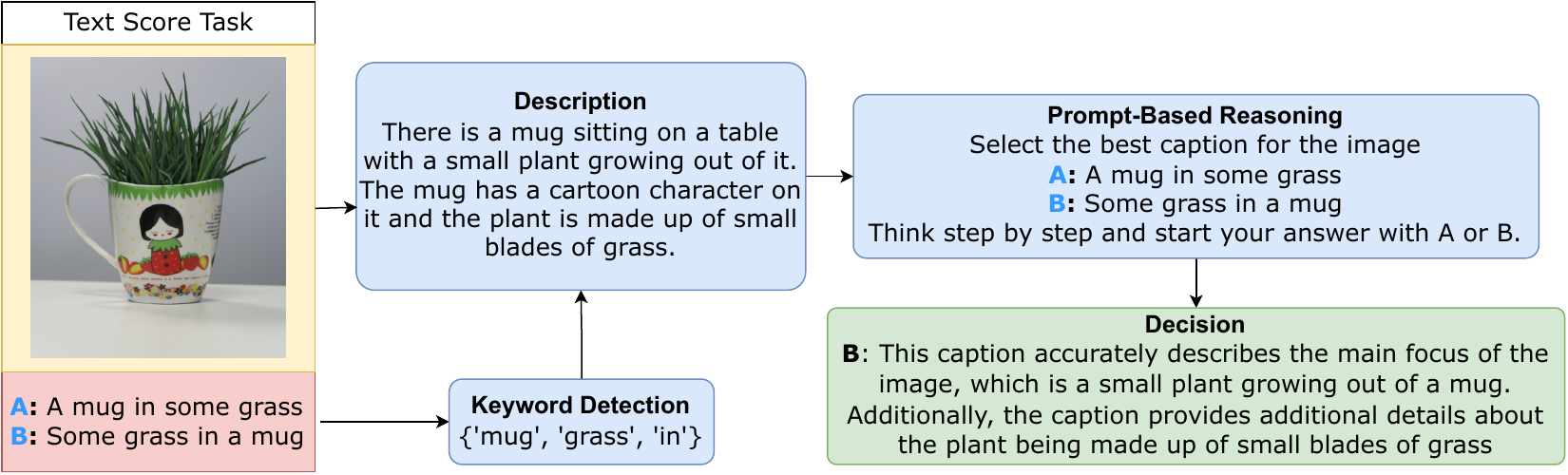}
  \caption{No Tag Question. }
  \label{fig:non_compositional}
\end{figure}

\clearpage
\label{sec:question_categories}

\section{Image Score Task for Question Categories}
In this section we illustrate the details of our method when applied to the image score task. We include one example from each question category, using the same image caption pairs used in Appendix \ref{sec:question_categories}.
\label{sec:image_score_examples}
\begin{figure}[!htb]
  \includegraphics[width=\textwidth]{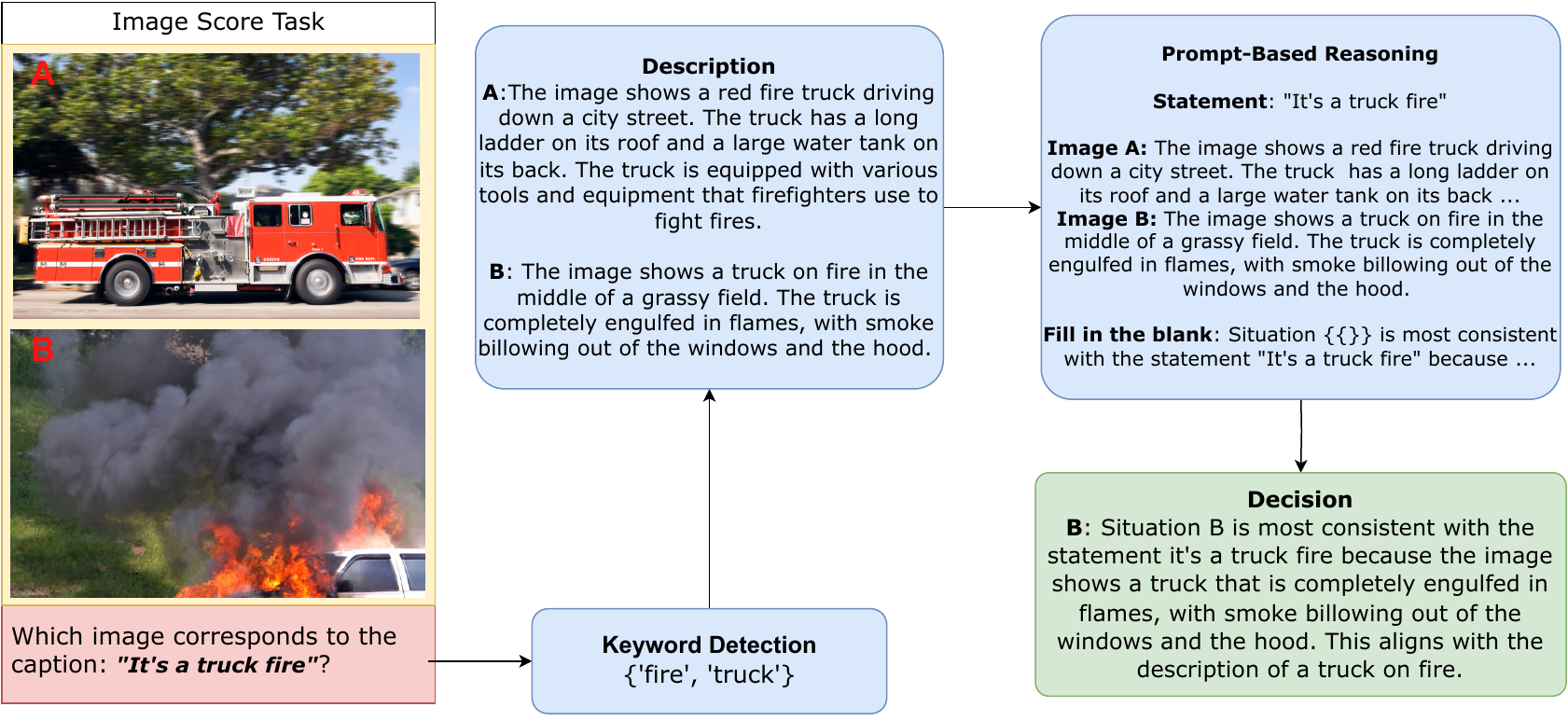}
  \caption{Non-Compositional Question for the image score task.}
  \label{fig:non_compositional}
\end{figure}
\begin{figure}[!htb]
  \includegraphics[width=\textwidth]{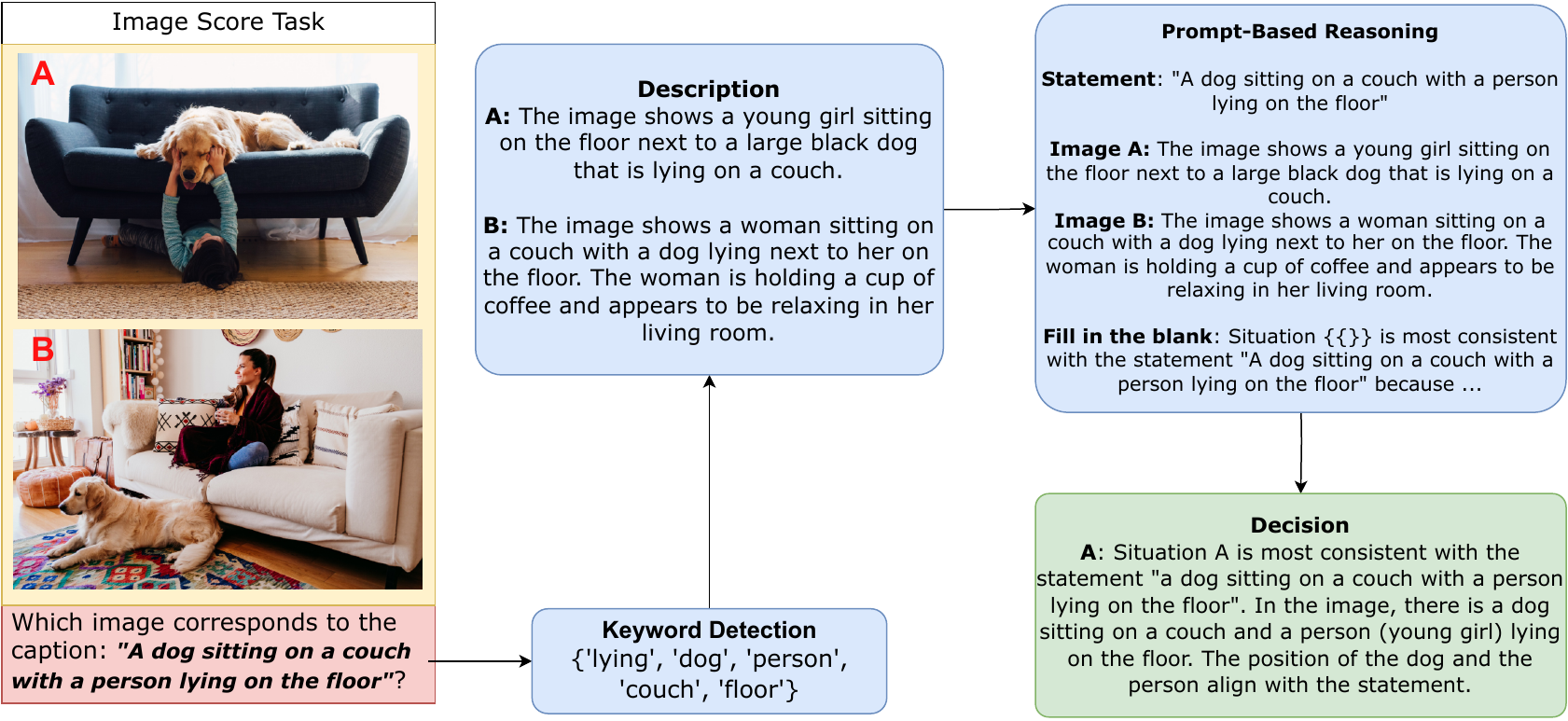}
  \caption{Ambiguously Correct Question. Refer to Appendix \ref{sec:question_categories} for the ambiguity contained in image B. }
  \label{fig:non_compositional}
\end{figure}

\begin{figure}[!htb]
  \includegraphics[width=\textwidth]{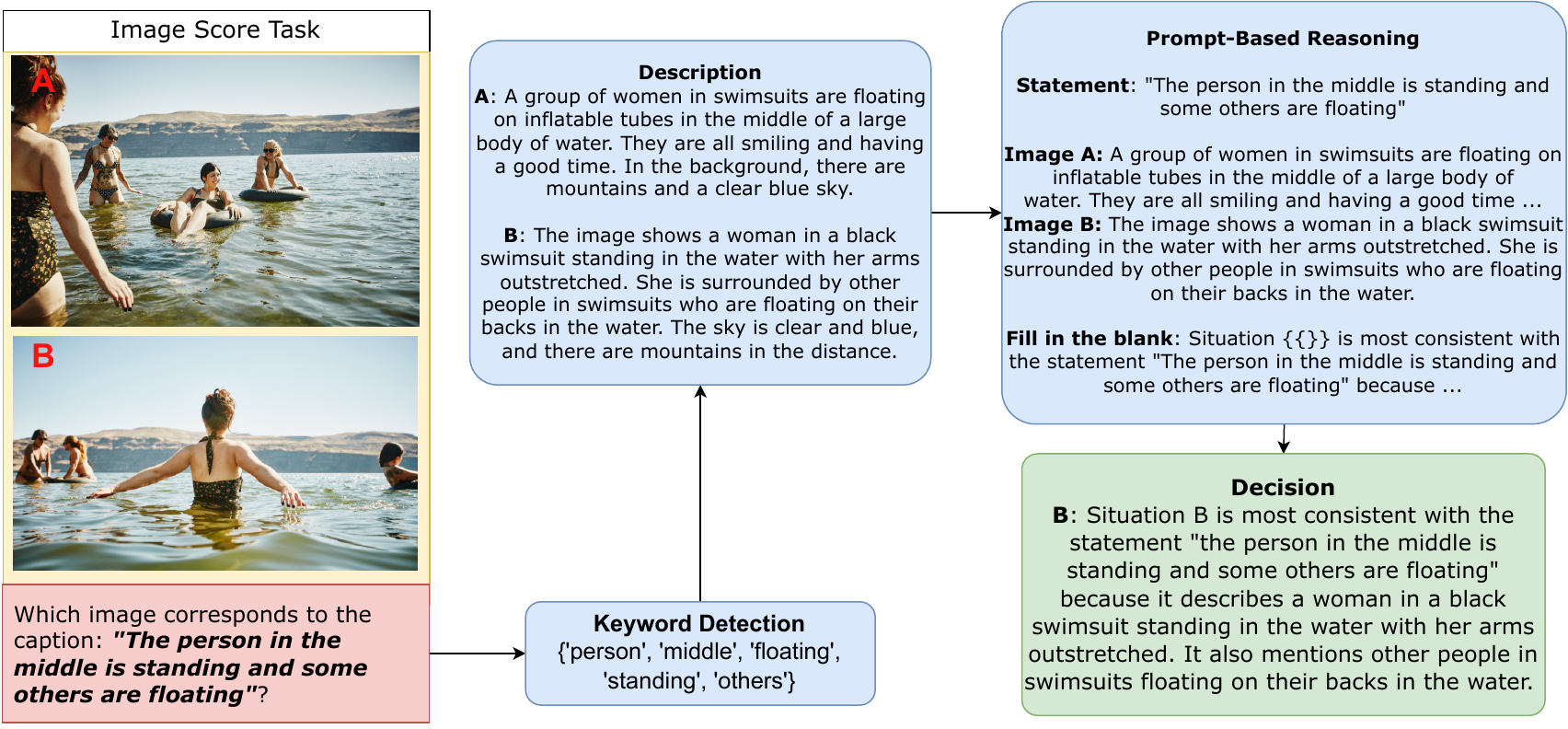}
  \caption{Visually Difficult Question for the image score task. }
  \label{fig:non_compositional}
\end{figure}

\begin{figure}[!htb]
  \includegraphics[width=\textwidth]{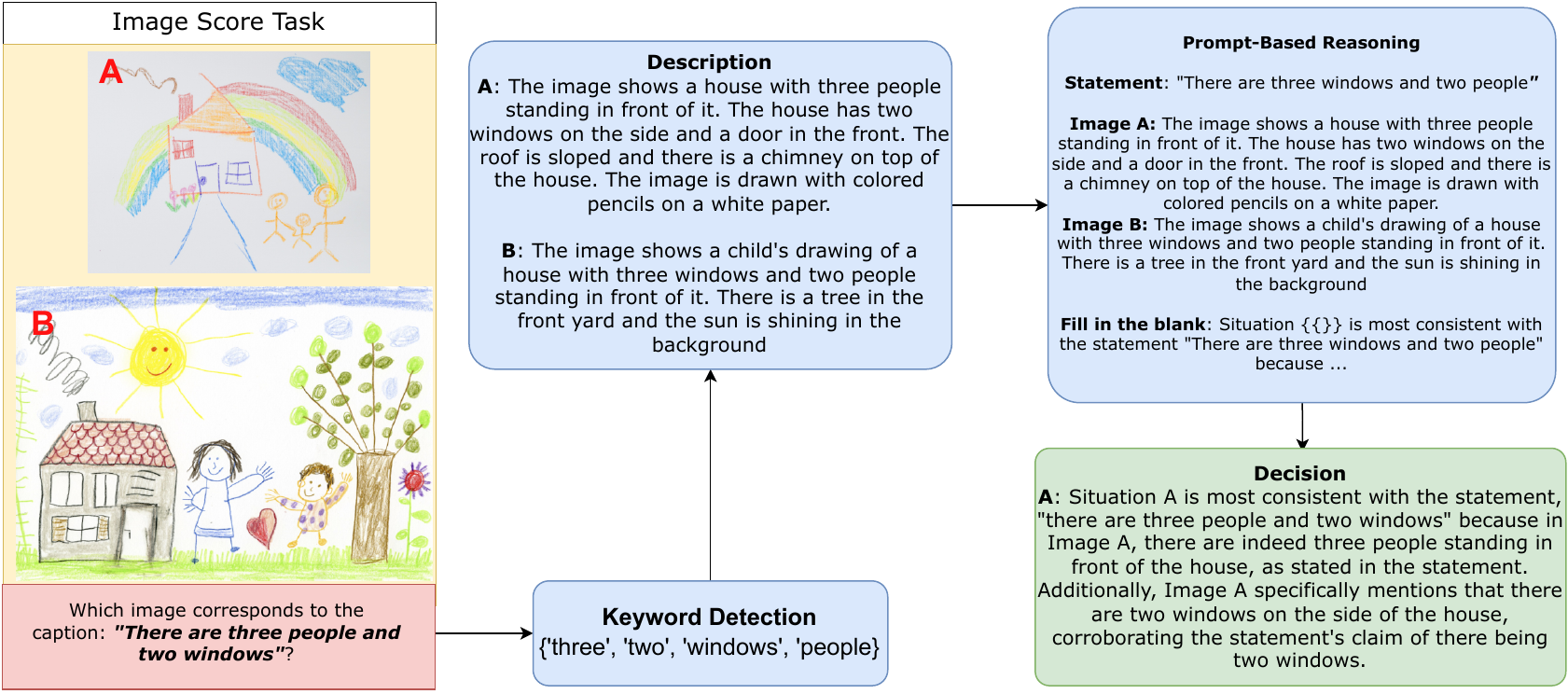}
  \caption{Unusual Image Question. Although both images are cartoons, both descriptions match the contents of the images accurately. }
  \label{fig:non_compositional}
\end{figure}
\begin{figure}[!htb]
  \includegraphics[width=\textwidth]{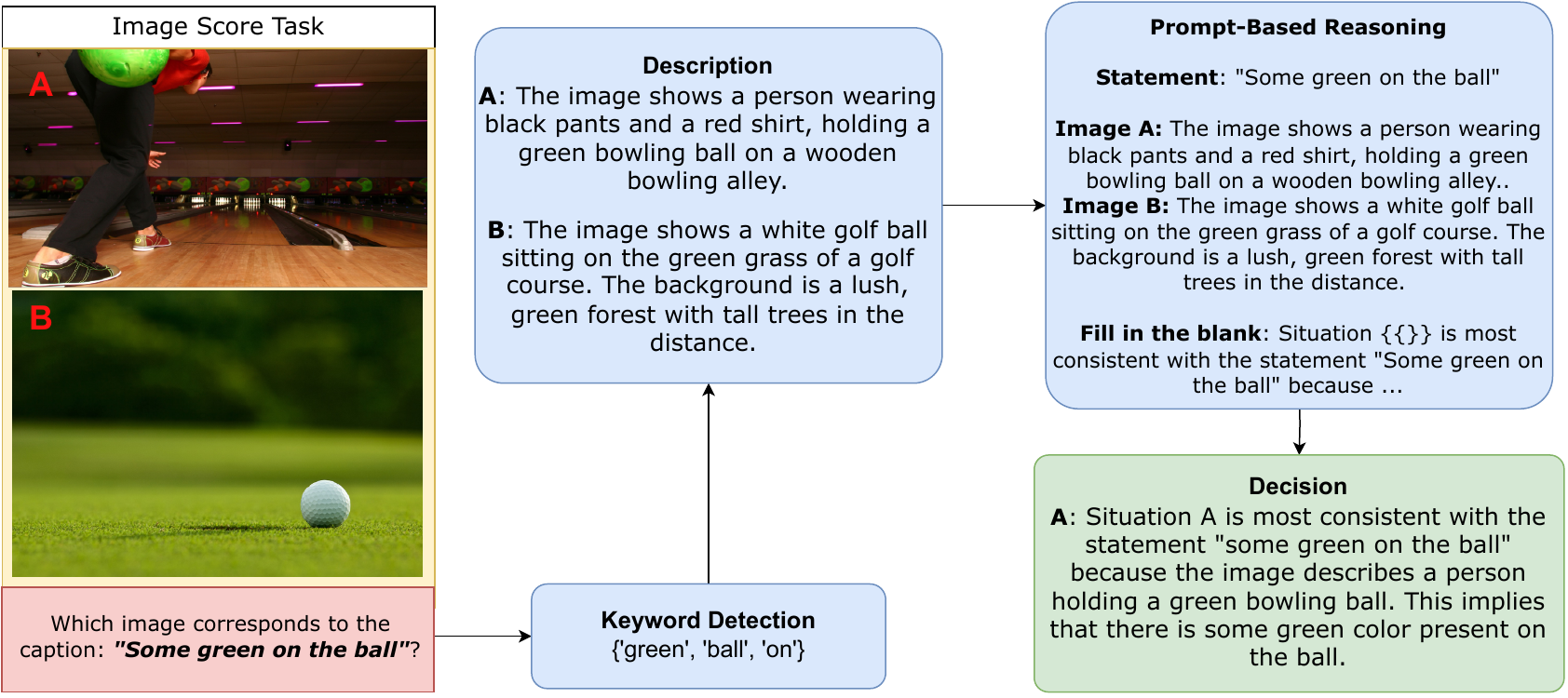}
  \caption{Unusual Text question for the image score task. }
  \label{fig:non_compositional}
\end{figure}
\begin{figure}[!htb]
  \includegraphics[width=\textwidth]{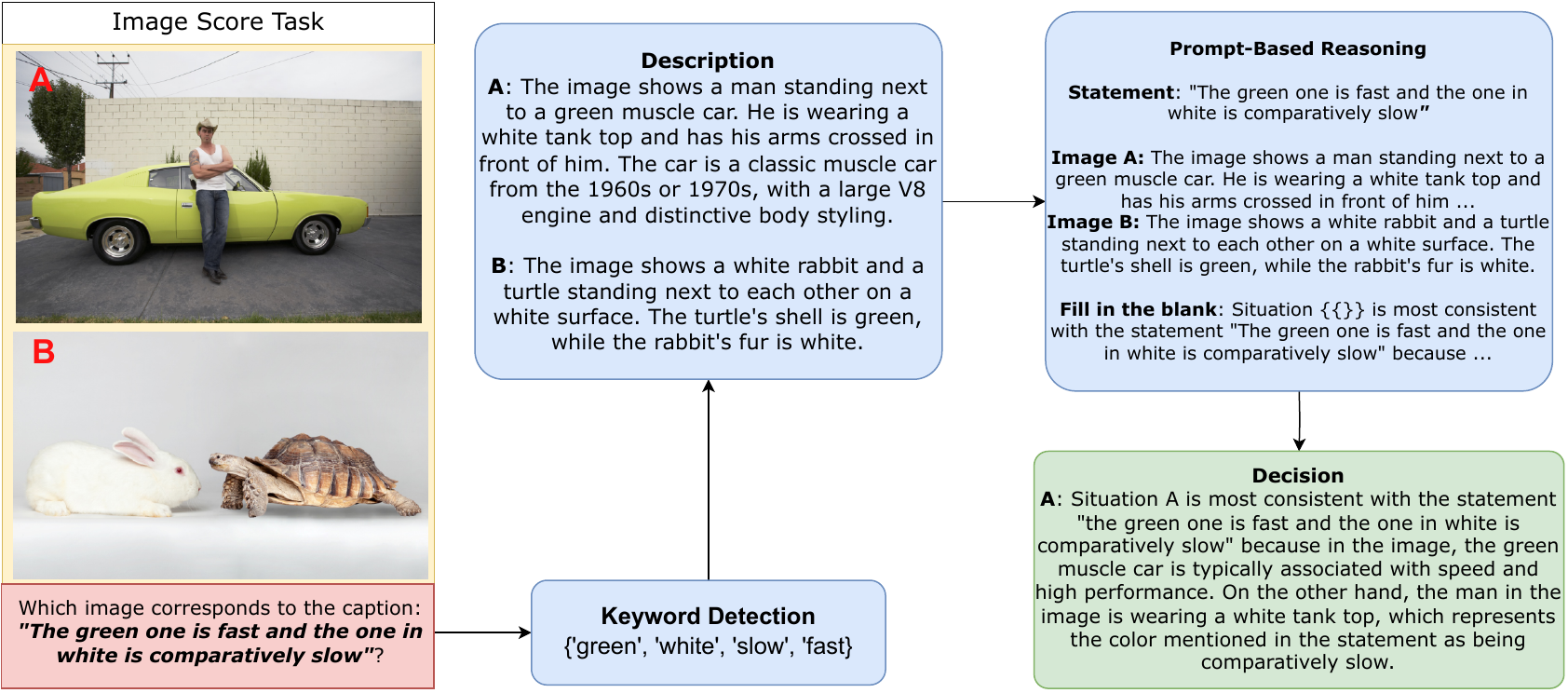}
  \caption{Complex Reasoning question for the image score task. }
  \label{fig:non_compositional}
\end{figure}
\begin{figure}[!htb]
  \includegraphics[width=\textwidth]{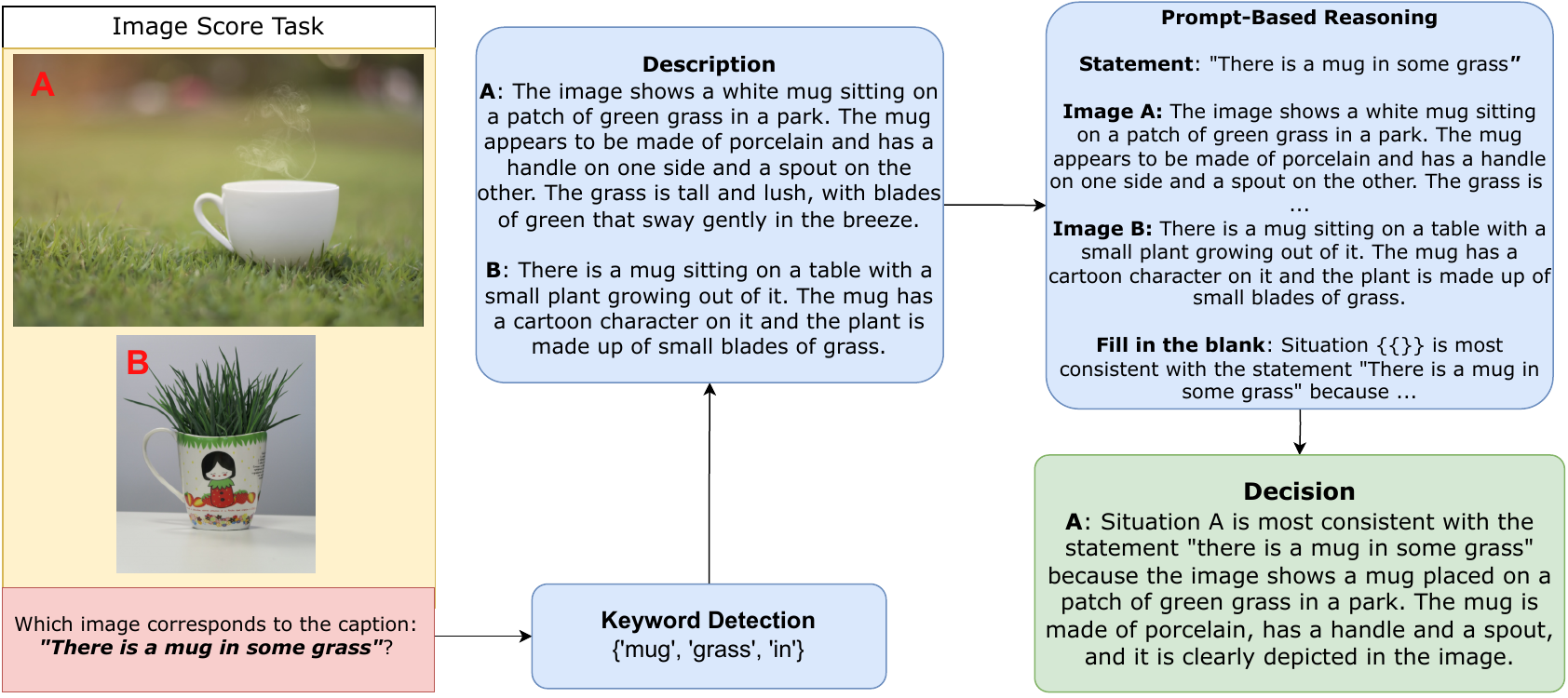}
  \caption{NoTag question for the image score task. }
  \label{fig:non_compositional}
\end{figure}

\clearpage

\section{Fine Grained Image Scores}

\begin{figure*}[!htb]
  \includegraphics[width=\textwidth]{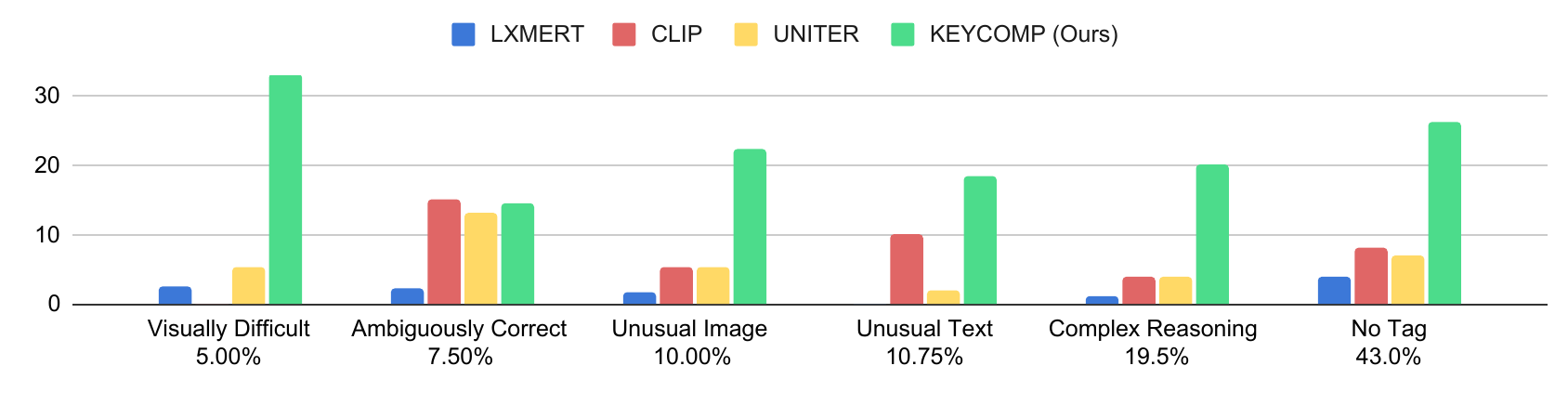}
  \vspace{-1cm}
  \caption{Fine-grained image score performance across different question categories. We give specific examples from each category in Appendix \ref{sec:question_categories}. Percentages on the x-axis indicate each question type's proportion of the dataset. To ensure representative results, question categories comprising less than 5\% of the dataset are excluded.}
  \label{fig:fine_grained_text}
\end{figure*}
\label{sec:fine_grained_text}

\section{Effect of VLM Size and LLM Size}
\label{sec:vlm_llm_size}
\input{tables/VLM_Size}
\input{tables/LLM_Size}

Tables \ref{tab:vlm_size} and \ref{tab:llm_size} show the effect of changing the VLM and LLM sizes respectively. Our results suggest that changing the LLM size improves text score significantly and offers marginal improvement in image score. This intuitively makes sense because text score requires a deeper understanding of text to distinguish between similar captions.  Upgrading the VLM also provides improvement to text score, but does not impact the image score.
\clearpage
\section{Error Analysis}
\label{sec:error_analysis}
In this section, we detail what examples \ours struggles with and suggest future directions for generative approaches. We categorize common errors as either VLM-based, LLM-based or both.

\subsection{Spatial Reasoning (VLM)}
For questions requiring spatial reasoning, we note that \ours sometimes produces inaccurate descriptions of the scene. We believe this stems from the image captioning model (VLM) and may be addressed by utilizing scene understanding models trained with object-relation level supervision.
\begin{figure*}[!htb]
  \includegraphics[width=\textwidth]{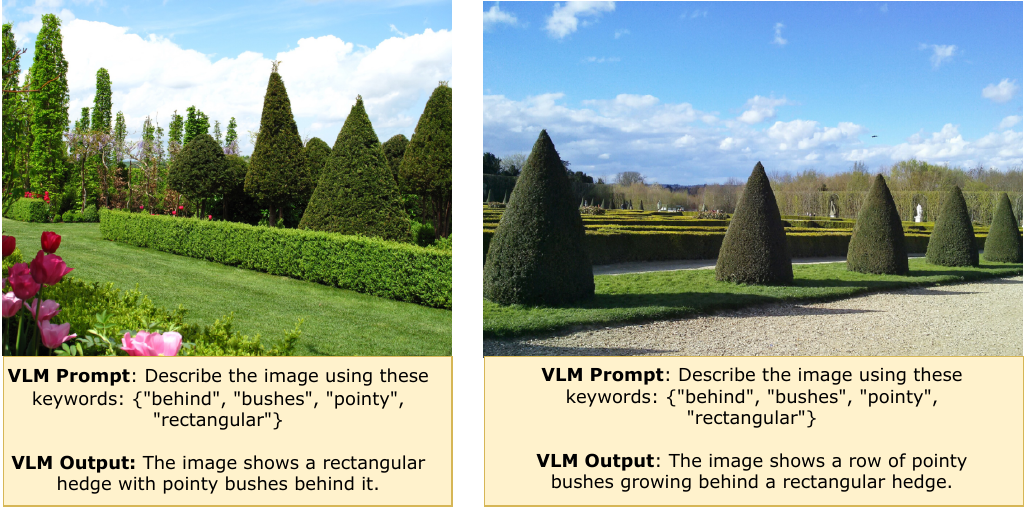}
  \caption{Spatial Reasoning Error. As the VLM does not explicitly model the spatial information of objects in an image, the VLM mistakenly thinks the bushes are behind the hedges in the right image. This is particularly difficult when two objects have similar colors.}
  \label{fig:spatial_error}
\end{figure*}
\subsection{Incorrect LLM Reasoning}
Occasionally the LLM is presented with VLM descriptions that should contain enough information to answer the question, but still selects the wrong answer. This often occurs when the VLM description has a complicated syntax structure that involves multiple objects. 

\begin{figure*}[!h]
\begin{center}
    \includegraphics[width=0.8\textwidth]{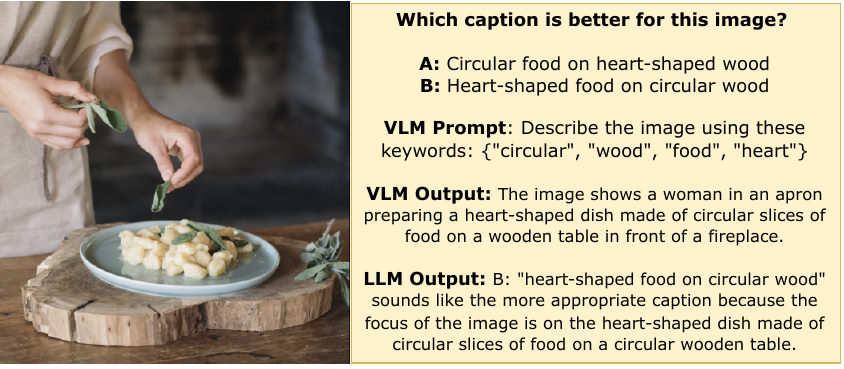}
\end{center}
  
  \caption{LLM Interpretation Error. Although VLMs provide a sufficiently detailed description, the description has a complicated syntax structure that involves multiple objects. The LLM response still does not match the VLM description.}
  \label{fig:llm_error}
\end{figure*}

\subsection{Describing Out of Focus or Occluded Objects (VLM)}
Sometimes the VLM produces inaccurate descriptions of scene elements which are out of focus or missing parts of objects. In these cases, even though the prompt requires the LLM to choose an option, the LLM may sometimes refuse to choose, saying it is uncertain and offers a better caption. This makes it hard to automate the evaluation and may be addressed with further fine-tuning in this instruction format or additional prompt engineering.

\begin{figure*}[!htb]
  \includegraphics[width=\textwidth]{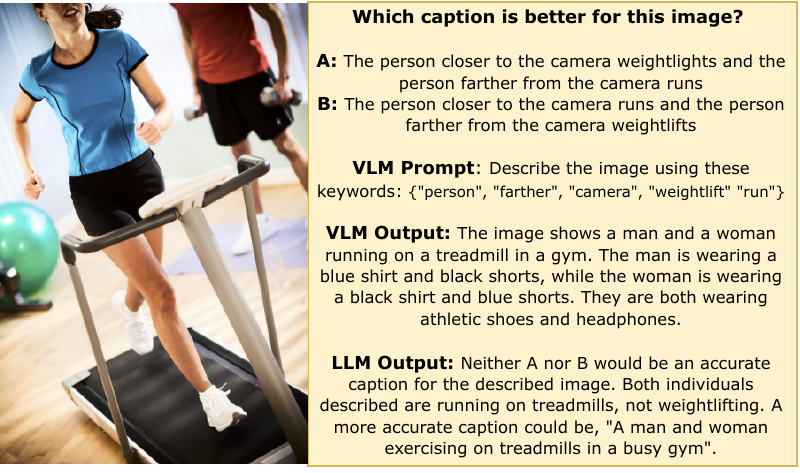}
  \caption{LLM does not give an answer. In this case, the VLM description also does not describe the scene accurately, likely because the man is out of focus and his head is not visible.}
  \label{fig:invisible_error}
\end{figure*}

%% file: tables/prompt_table_text.tex
\begin{table}[!htb]
    \centering
    
    \resizebox{\linewidth}{!}{
    \begin{tabular}{c|c|c|c}
    \toprule[1pt]
        Name & VLM Prompt ($P_\mathcal{K}$) & ChatGPT Prompt ($P_{\text{txt}, a}$) & Performance \\
        \midrule
        \ours & \makecell{Describe the image in detail \\ using these keywords: \texttt{\{keywords\}}} & \makecell{Select the best caption for this image:  \\  A: \texttt{\{caption\_0\}} \\ B: \texttt{\{caption\_1\}} \\ Think step-by-step and start your answer with A or B. \\ Even if you are unsure make a guess. } & 30.3 \\
        \midrule
        Explanation Prompting & \makecell{Describe the image in detail \\ using these keywords: \texttt{\{keywords\}}} & \makecell{Select the best caption for this image:  \\  A: \texttt{\{caption\_0\}} \\ B: \texttt{\{caption\_1\}} \\ Start your answer with A or B. \\ Even if you are unsure make a guess. \\ Briefly explain your decision in 1-2 sentences.} & 26.3 \\
        \midrule
        
        Multiple Choice & \makecell{Describe the image in detail \\ using these keywords: \texttt{\{keywords\}}} & \makecell{Select the best caption for this image:  \\  A: \texttt{\{caption\_0\}} \\ B: \texttt{\{caption\_1\}} \\ Start your answer with A or B. \\ Even if you are unsure make a guess. } & 29.0 \\
        \midrule
        No Keyword Guidance & \makecell{Describe the image in detail} & \makecell{Select the best caption for this image:  \\  A: \texttt{\{caption\_0\}} \\ B: \texttt{\{caption\_1\}} \\ Start your answer with A or B. \\ Even if you are unsure make a guess. \\ Briefly explain your decision in 1-2 sentences.} & 21.6 \\
        \midrule
        MiniGPT4 End-to-End & \makecell{Which caption is more appropriate:  \\  A: \{\texttt{caption\_0}\} \\ B: \{\texttt{caption\_1}\} \\ Answer in one sentence. \\ Even if you are unsure make a guess. \\ Briefly explain your decision in 1-2 sentences.} & None & 20.8 \\

    \bottomrule[1pt]
    \end{tabular}
    
    }
    
    \vspace{2mm}
    \caption{Average performance of different prompt variants for text score across 3 different runs.}
    \label{tab:prompts}
\end{table}

%% file: tables/prompt_table_image_score.tex
\begin{table}[!htb]
    \centering
    
    \resizebox{\linewidth}{!}{
    \begin{tabular}{c|c|c|c}
    \toprule[1pt]
        Name & VLM Prompt ($P_\mathcal{K}$) & ChatGPT Prompt ($P_{\text{img}, a}$) & Performance \\
        \midrule
        \ours & \makecell{Describe the image in detail \\ using these keywords: \{\texttt{keywords}\}} & \makecell{Statement: \{\texttt{caption}\} \\  Image A: \{\texttt{description\_0}\} \\ Image B: \{\texttt{description\_1}\} \\ Think step by step and fill in the blank: \\ Situation \{\{\}\} is most consistent with the statement  \{\texttt{caption}\} because ... } & 24.6 \\
        \midrule
        Explanation Prompting & \makecell{Describe the image in detail \\ using these keywords: \{\texttt{keywords}\}} & \makecell{Statement: \{\texttt{caption}\} \\  Image A: \{\texttt{description\_0}\} \\ Image B: \{\texttt{description\_1}\} \\ Fill in the blank: \\ Situation \{\{\}\} is most consistent with the statement  \{\texttt{caption}\} because ... \\ Explain your decision in 1-2 sentences.} & 23.9 \\
        \midrule 
        Multiple Choice & \makecell{Describe the image in detail \\ using these keywords: \{\texttt{keywords}\}} & \makecell{Image 1: \{\texttt{image\_0\_description}\} \\  Image 2: \{\texttt{image\_1\_description}\} \\ Consider the caption: \{\texttt{caption}\} \\ Select the better image for this caption: \\ A: \{\texttt{image\_1}\} \\ B: \{\texttt{image\_2}\} \\ Start your answer with A or B. } & 22.3 \\
        \midrule
        No Keyword Guidance & \makecell{Describe the image in detail} & \makecell{Statement: \{\texttt{caption}\} \\  Image A: \{\texttt{description\_0}\} \\ Image B: \{\texttt{description\_1}\} \\ Think step by step and fill in the blank: \\ Situation \{\{\}\} is most consistent  with the statement  \{\texttt{caption}\} because ... } & 21.8 \\
        \midrule
        MiniGPT4 End-to-End & \makecell{Given the following images: \\ A: \texttt{<Img>ImageContent</Img>} \\ B: \texttt{<Img>ImageContent</Img>}  \\  
        Which image is more appropriate \\ for the caption \texttt{\{caption\}}? \\ Answer with A or B.} & None & 12.5 \\

    \bottomrule[1pt]
    \end{tabular}
    
    }
    
    \vspace{2mm}
    \caption{Average performance of different prompt variants for image score across 3 separate runs.}
    \label{tab:prompts_image}
\end{table}

%% file: tables/VLM_Size.tex
\begin{table}[thb!]
    \centering
    
    \resizebox{7cm}{!}{
    \begin{tabular}{l l l l}
    \toprule[1pt]
        VLM & Text & Image & Group \\
        \midrule
         MiniGPT-4 7b & 23.3 & 25.5  & 0.1  \\
         MiniGPT-4 13b & 30.3 &  24.6  & 12.4  \\

    \bottomrule[1pt]
    \end{tabular}
    }
    \caption{\ours performance while changing the size of the VLM image captioner.}
    \label{tab:vlm_size}
    \vspace{-3mm}
\end{table}

%% file: tables/LLM_Size.tex
\begin{table}[thb!]
    \centering
    
    \resizebox{6cm}{!}{
    \begin{tabular}{l l l l}
    \toprule[1pt]
        LLM & Text & Image & Group \\
        \midrule
        GPT-3.5 & 30.3 &  24.6  & 12.4  \\
        GPT-4 & 44.0 & 25.9  & 15.4  \\

    \bottomrule[1pt]
    \end{tabular}
    }
    \caption{\ours performance while changing the size of the LLM used for reasoning.}
    \label{tab:llm_size}
    \vspace{-3mm}
\end{table}